\documentclass[letterpaper, 10 pt, conference]{ieeeconf}

\IEEEoverridecommandlockouts
\usepackage{amsmath,amssymb}
\usepackage{graphicx}
\usepackage{booktabs}
\usepackage{cite}
\usepackage{url}
\usepackage{dblfloatfix}

\title{\LARGE \bf
Emo-H2A: Driving a Facial Robot Directly from LLM Hidden States at Token Rate
}

\author{Yige Liu$^{1}$, Jingcheng Huang$^{1}$, Hao Sun$^{1}$, Min~Li$^{1,*}$, and Zhouping Yin$^{1}$%
\thanks{$^{*}$Corresponding author: Min Li ({\tt\small min.li@hust.edu.cn}).}%
\thanks{$^{1}$Huazhong University of Science and Technology, Wuhan 430074, China.
Emails: {\tt\small yigeliu@hust.edu.cn}, {\tt\small D202380251@hust.edu.cn},
{\tt\small m202670455@hust.edu.cn}, {\tt\small yinzhp@mail.hust.edu.cn}.}%
}

\begin{document}

\maketitle
\thispagestyle{empty}
\pagestyle{empty}

\begin{abstract}
Diverse, semantically synchronized, and low-latency facial expressions are essential for human-robot interaction. While large language models (LLMs) have endowed anthropomorphic robots with fluent speech, facial expression still struggles to follow the conversation. Motivated by this gap, we present Emo-H2A (hidden-to-actuation), a paradigm that drives a facial robot directly from LLM hidden states during autoregressive decoding. Without a separate emotion label, expression stays continuous and context-sensitive, and the face moves with the unfolding reply. We instantiate this paradigm as EmoManifold, a training-free method that extracts a valence--arousal--dominance (VAD) coordinate from each hidden state, projects it, and drives the face at token granularity. The readout is checked on six language models, and the same projection drives our 29-degree-of-freedom (DoF) facial robot. When driven by this projection, the face shifts its expression in the matching direction on $87.5\%$ of within-reply valence transitions---whereas holding a single label for the whole reply leaves directional tracking near the $50\%$ chance level---and first moves in $0.96$~s rather than waiting $10$--$26$~s for the reply to finish and a turn-level label to return.
\end{abstract}

\begin{figure*}[t]
    \centering
    \includegraphics[width=\textwidth,keepaspectratio]{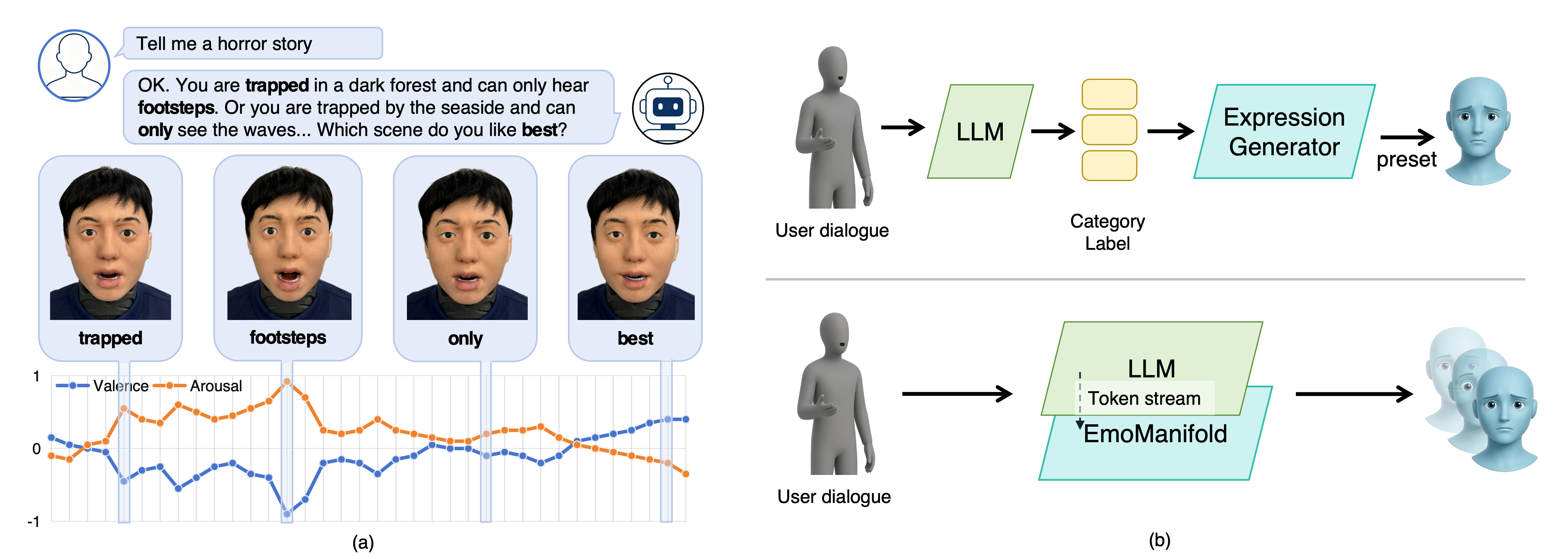}
    \caption{\textbf{Emo-H2A.} \textbf{(a)}~Schematic of token-level expression. As each token is generated, continuous valence and arousal coordinates are read out (bottom), driving the facial robot to transition smoothly across nuanced expressions (shown at selected tokens). \textbf{(b)}~\textbf{Overview.} Existing pipelines (above) pass the reply through a category label and an expression generator to play one preset~\cite{mishra2023realtime,antony2025xpress}. EmoManifold (below) drives the face directly from the token stream as the reply is generated.}
    \label{fig:effect}
\end{figure*}

\begin{figure*}[t]
    \centering
    \includegraphics[width=\textwidth,keepaspectratio]{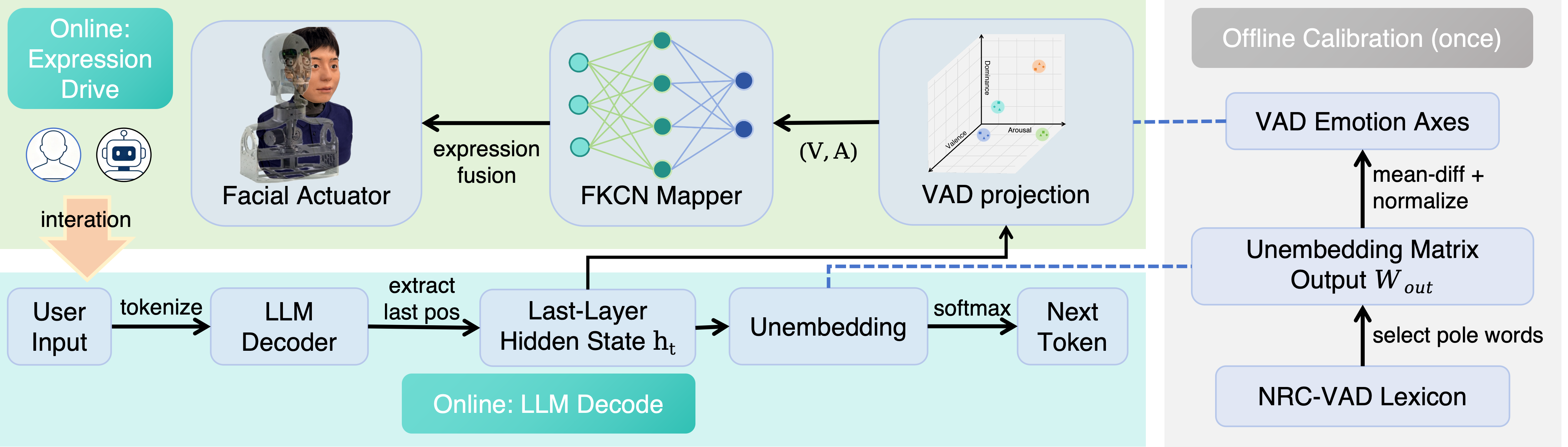}
    \caption{\textbf{EmoManifold.} Online decoding (bottom): user input is tokenized and passed through the LLM decoder; the final-layer, last-position hidden state $\mathbf{h}_t$ is extracted and unembedded to obtain the next token. Online expression driving (top): $\mathbf{h}_t$ is projected onto VAD; its valence--arousal (VA) components are mapped by a Fuzzy Kohonen Clustering Network (FKCN) into seven-dimensional expression weights, fused with presets to drive the facial actuators. Offline calibration (right): NRC-VAD pole words and the unembedding matrix $\mathbf{W}_{\mathrm{out}}$ construct the affective axes once, offline and training-free.}
    \label{fig:system}
\end{figure*}

\section{INTRODUCTION}

In face-to-face communication, humans perceive spoken words and facial expressions as an inseparable composite. To reproduce this natural affinity in embodied systems, anthropomorphic platforms with high-DoF physical actuation are actively being developed~\cite{hu2024emo,li2025x2c}, while modern LLMs provide the linguistic engine for fluent dialogue. Yet, despite these advances, modern conversational robots remain fundamentally desynchronized: the facial robot falls out of step with the affective nuance of its own speech.

This desynchronization stems from treating emotion as an extrinsic label rather than an endogenous signal. In existing pipelines, an utterance opening with sympathy and closing with encouragement is assigned a single categorical tag, appearing only after the full reply has been generated~\cite{mishra2023realtime,cherakara2023furchat}. Recent prompt-engineering and multi-stage scheduling approaches attempt to develop intra-utterance trajectories by matching generated segments to expression presets~\cite{antony2025xpress}. However, this strategy introduces complex prompt designs and additional parsing passes that substantially increase inference latency. As a result, conventional frameworks struggle to balance response latency against affective granularity, highlighting the need to move beyond label-mediated emotion generation.

Affective diversity presents another major hurdle. Traditional discrete emotion categories correspond to only a handful of static motor templates, which fail to match the rich and nuanced expressive capabilities of modern language models. Consequently, an urgent need exists for a continuous affective space capable of modeling fine-grained emotional transitions. Recent interpretability studies find a low-dimensional affective manifold in LLM hidden states, with directions resembling valence, dominance, and arousal~\cite{reichman2026emotions}. The emergence of this VAD-like geometry suggests that LLMs can provide the representational foundation for continuous facial affect modeling.

To address these challenges, we introduce \textbf{Emo-H2A}, a paradigm that couples the LLM's autoregressive hidden states at each generated token directly to facial actuator commands. We instantiate this paradigm as \textbf{EmoManifold}, an online framework that projects token-level hidden representations onto a continuous VAD subspace derived from the model's unembedding vectors and maps the resulting affective coordinates to physical motor commands. Physical deployment on our 29-DoF facial robot demonstrates that EmoManifold follows within-utterance affective shifts with low initial response latency, achieving an $83.5\%$ preference rate from human observers over both a static expression held for the entire utterance and a temporally shuffled control.

To foster further research, we will open-source our codebase and evaluation benchmarks. In summary, our main contributions are:
\begin{itemize}
\item \textbf{Emo-H2A}: a paradigm that maps the hidden state at each decoding step to facial motor commands, without a separate emotion label.
\item \textbf{EmoManifold}: a training-free method that builds a VAD subspace, projects each token's hidden state onto it, and drives the face from that coordinate.
\item \textbf{Physical evaluation}: On our 29-DoF facial robot, token-level actuation tracks within-reply transitions above chance, where turn-level baselines stay near chance, and observers prefer the token-level motion over a static expression.
\end{itemize}

\section{RELATED WORK}
\label{sec:related}

\subsection{Facial Robot Platforms and Expression Generation}
The evolution of expressive facial robots has progressed from simplified, rule-driven mechanisms to high-DoF platforms coupled with LLMs. Early facial robots featured relatively simple mechanical linkages---such as Kismet~\cite{breazeal2002designing}, which actuated eyes, eyebrows, ears, and lips---and synthesized expressions through hand-crafted rules grounded in the Facial Action Coding System (FACS)~\cite{ekman1978facs} or prototype interpolation. Subsequent advances in compliant skin materials and compact multi-actuator integration enabled high-DoF facial robots capable of nuanced surface deformation. Within this generation, data-driven visual imitation became prominent: Emo~\cite{hu2024emo} trains self-supervised vision-to-motor models on a 26-actuator facial robot to anticipate and coexpress human facial movements, while X2C~\cite{li2025x2c} maps observed human faces onto 30 continuous actuator channels on the Ameca anthropomorphic robot. Although these platforms achieve lifelike mechanical articulation, their control signals are driven by external visual mimicry rather than conversational semantics. To align nonverbal behavior with spoken dialogue on anthropomorphic robots, recent systems integrate LLMs into the expression pipeline. FurChat~\cite{cherakara2023furchat}, SAFE~\cite{lee2023safe}, Nadine~\cite{kang2024nadine}, and Mishra et al.~\cite{mishra2023realtime} query or prompt LLMs to attach discrete emotion tags or nonverbal cue bundles to completed replies, whereas Expressive Furhat~\cite{abbo2025expressive} and Xpress~\cite{antony2025xpress} generate finer-grained gesture tags, intensity levels, or keyframe schedules via structured prompting. However, all existing LLM-driven pipelines treat expression as an extrinsic symbolic output generated or parsed after text decoding, whereas Emo-H2A treats affect as an endogenous signal read out at each decoding step.

\subsection{Affective Representations in Expressive Robotics}
How affective states are represented fundamentally determines the granularity and smoothness of robotic facial control. Historically, social robotics has relied predominantly on discrete categorical models---such as Ekman's basic emotion categories~\cite{ekman1978facs}---to index static motor presets or turn-level dialogue classifiers~\cite{mishra2023realtime,cherakara2023furchat,kang2024nadine}. While straightforward to calibrate on hardware, discrete labels impose coarse quantization and abrupt transitions that cannot capture subtle within-utterance emotional shifts. To enable continuous modulation, expressive robotics turned to dimensional affect models from psychology, notably Russell's two-dimensional circumplex model of valence and arousal~\cite{russell1980circumplex} and Russell and Mehrabian's three-factor VAD space~\cite{russell1977evidence}, later operationalized at scale by the NRC-VAD lexicon~\cite{mohammad2018nrcvad}. Early platforms such as Kismet~\cite{breazeal2002designing} interpolated postures within a designer-specified three-dimensional space of arousal, valence, and stance. Yet in contemporary LLM-based robots, continuous affective coordinates still rely on post-hoc classifiers or explicit scalar prompting~\cite{antony2025xpress}. Parallel mechanistic interpretability research reveals that LLMs inherently encode linear affective geometry within their latent spaces: Tigges et al.~\cite{tigges2024sentiment} extract a linear sentiment direction from residual-stream activations using contrastive sentence pairs, and Reichman et al.~\cite{reichman2026emotions} uncover a cross-layer emotional manifold that allows latent emotion steering without altering surface semantics. Whereas prior interpretability studies recover affective directions by running forward passes over labeled or contrastive corpora or by training auxiliary modules, EmoManifold reads its VAD axes from the unembedding matrix, without training an auxiliary module.

\section{METHOD}
\label{sec:method}

The face changes with the tokens of a reply, instead of waiting for one label (Fig.~\ref{fig:effect}), and EmoManifold carries each decoding step to a facial command before the reply is finished (Fig.~\ref{fig:system}). This section first constructs the affective axes (Section~\ref{sec:axis_mining}); next, it projects each token's hidden state onto those axes (Section~\ref{sec:projection}); finally, it maps the coordinates to actuator commands (Section~\ref{sec:actuator}). The implementation follows three principles:
\begin{itemize}
\item \textbf{Expression-Rich}: Use a continuous space of expressions to capture slight affective changes.
\item \textbf{Token-Level}: Update the face at a token rate.
\item \textbf{Direct-Readout}: Derive the facial command from one projection, and introduce no reasoning step that would add latency.
\end{itemize}

\subsection{Training-Free Affective Axes}
\label{sec:axis_mining}

This stage builds three affective axes in the LLM's hidden space, aligned with the VAD model~\cite{russell1977evidence}. Each axis is a difference of unembedding rows, so no emotion classifier is trained and the language model is not fine-tuned.
Let $\mathbf{W}_{\mathrm{out}} \in \mathbb{R}^{|\mathcal{V}| \times d}$ denote the model's unembedding matrix, where $|\mathcal{V}|$ is the vocabulary size and $d$ is the hidden dimension.
For each dimension $k \in \{V, A, D\}$, we form positive and negative pole word clusters $\mathcal{C}^{+}_k$ and $\mathcal{C}^{-}_k$ from the NRC-VAD lexicon~\cite{mohammad2018nrcvad}. Words are ranked by extremity on that dimension, restricted to those that the target tokenizer encodes as a single token, and truncated to the top $N$ words per pole.
For a polarity cluster $\mathcal{C}^{*}_k$ ($* \in \{+, -\}$), we average the corresponding rows of the unembedding matrix:
\begin{equation}
    \bar{\mathbf{e}}_{\mathcal{C}^{*}_k} = \frac{1}{|\mathcal{C}^{*}_k|}\sum_{w \in \mathcal{C}^{*}_k} \bigl(\mathbf{W}_{\mathrm{out}}\bigr)_{\mathrm{id}(w)},
\end{equation}
where $\mathrm{id}(w)$ is the unique token index of word $w$.
The affective axis for dimension $k$ is the normalized difference of the two pole centroids:
\begin{equation}
    \mathbf{v}_k = \frac{\bar{\mathbf{e}}_{\mathcal{C}^{+}_k} - \bar{\mathbf{e}}_{\mathcal{C}^{-}_k}}{\bigl\|\bar{\mathbf{e}}_{\mathcal{C}^{+}_k} - \bar{\mathbf{e}}_{\mathcal{C}^{-}_k}\bigr\|_2}.
    \label{eq:valence_axis}
\end{equation}
The axes $\{\mathbf{v}_V, \mathbf{v}_A, \mathbf{v}_D\}$ span the VAD affective subspace. We use ``affective axis'' for per-dimension projection and construction, and ``affective subspace'' when discussing trajectory geometry, orthogonality, and expression mapping.
The pole-word count $N$ (default $50$; sensitivity analysis in Section~\ref{sec:exp_subspace}) is the main hyperparameter of axis construction.
The axes fix the directions, but the scale of a projection still depends on the model. A one-time affine map, AdaptiveScaler, is therefore fit on a small set of lexicon-word sentences, yielding per-dimension parameters $(s_k, b_k)$ that place projections in the lexicon's $[-1, 1]$ range. Those sentences are used once, offline, and the parameters stay fixed at inference.

\subsection{Per-Token VAD Projection}
\label{sec:projection}

At inference, the model generates tokens by causal decoding.
At step $t$, let $\mathbf{h}_t \in \mathbb{R}^{d}$ be the final-layer, last-position hidden state---the contextual representation of the token just emitted.
We L2-normalize $\mathbf{h}_t$ to remove norm drift across positions, project onto the three affective axes, and apply calibration:
\begin{equation}
\begin{gathered}
    \hat{\mathbf{h}}_t = \frac{\mathbf{h}_t}{\|\mathbf{h}_t\|_2}, \\[2pt]
    \tilde{z}_{t,k} = \mathrm{clip}\bigl(s_k\,\mathbf{v}_k^\top\hat{\mathbf{h}}_t + b_k,\,-1,\,1\bigr),
    \quad k \in \{V, A, D\}.
\end{gathered}
\label{eq:projection}
\end{equation}
where $\mathrm{clip}(\cdot,-1,1)$ saturates each coordinate to the lexicon range $[-1,1]$, and $\tilde{\mathbf{z}}_t=(\tilde{z}_{t,V},\tilde{z}_{t,A},\tilde{z}_{t,D})$. This produces a continuous VAD trajectory $\{\tilde{\mathbf{z}}_t\}$ at every decoding step, including the first token.
The projection adds three dot products to a hidden state the forward pass has already computed. On the deployment PC, the control path from this projection through expression blending and smoothing takes $0.7$~ms per token and does not include the language-model forward pass.

\subsection{Continuous Facial Mapping}
\label{sec:actuator}

This stage converts the VAD coordinates from~\eqref{eq:projection} into facial actuator commands in two steps. First, the coordinates are mapped through a basic-expression space into blend weights (expression mapping). Second, the weights are fused into servo commands with hardware-oriented smoothing and rate limiting (hardware actuation).

\subsubsection{Continuous Mapping from VAD to Basic Expressions}
Calibrating each servo directly in the VAD subspace is costly, and the resulting map is tied to one joint layout. The basic-expression space provides a more practical calibration basis.
We therefore use an FKCN~\cite{han2010fkcn} to map the VA components into blend weights $\boldsymbol{\omega}_t \in \Delta^{6}\subset\mathbb{R}^{7}$ (nonnegative and summing to one) over seven basic expressions (neutral, happiness, surprise, fear, sadness, disgust, and anger).
The full VAD coordinate is retained in the readout, and each dimension is evaluated in Section~\ref{sec:experiments}. For facial actuation, we use only valence and arousal: dominance accounts for less variance in affective judgments than valence or arousal~\cite{bradley1994measuring}, draws the lowest inter-rater agreement among the three when observers rate a face~\cite{sutton2019valence}, and is classically carried by posture~\cite{mehrabian1969significance}. Because the coordinate already captures dominance, a downstream mapping can route it to the actuators without rebuilding the axes; in this pipeline, it remains in the readout rather than the servo command.

\begin{figure}[t]
    \centering
    \includegraphics[width=0.82\columnwidth,keepaspectratio]{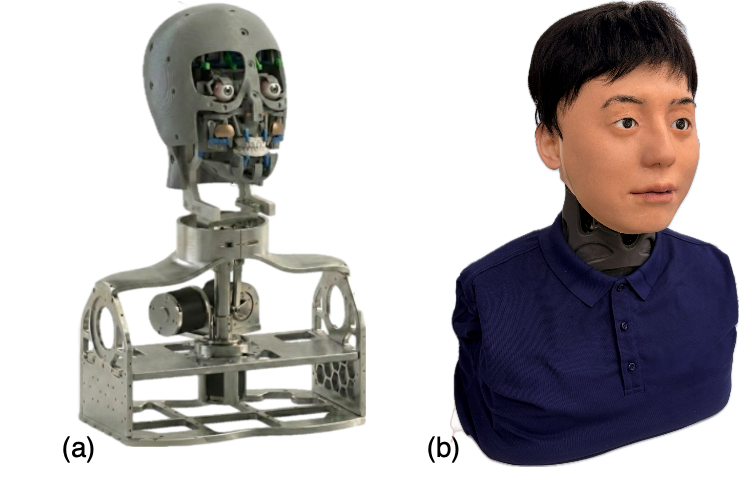}
    \caption{The 29-DoF facial robot platform used for the physical evaluation. \textbf{(a)}~Unskinned mechanical skeleton: servo groups span the eyelids, brows, and mouth. \textbf{(b)}~Full appearance with simulated silicone skin---the configuration driven and recorded in the experiments.}
    \label{fig:add}
\end{figure}

\begin{figure*}[t]
    \centering
    \includegraphics[width=\textwidth,keepaspectratio]{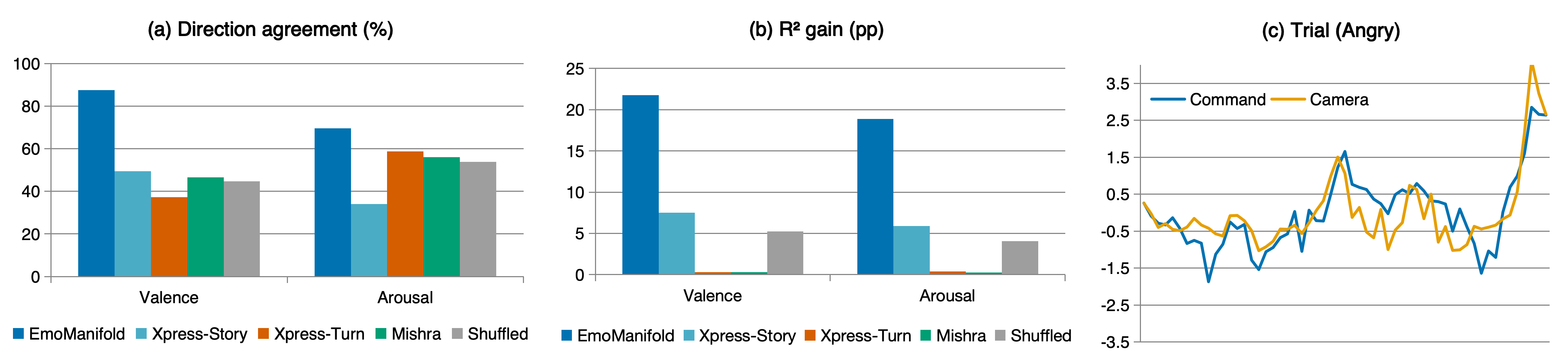}
    \caption{Token-level affective physical actuation ($n{=}20$ replies). Condition definitions are in Section~\ref{sec:exp_hardware}.
    \textbf{(a)}~Directional agreement between commanded affective transitions and the camera readout.
    \textbf{(b)}~$R^2$ gain, in percentage points, beyond a turn-level expression.
    \textbf{(c)}~EmoManifold on the representative anger trial: command (blue) and calibrated camera (amber), $z$-scored valence.}
    \label{fig:5}
\end{figure*}

The FKCN has $P = 13$ fuzzy prototypes $\{(\mathbf{c}_j, \mathbf{w}_j)\}_{j=1}^{P}$, where $\mathbf{c}_j \in \mathbb{R}^2$ is a VA-plane coordinate and $\mathbf{w}_j\in\{0,1\}^{7}$ is the one-hot label of the associated basic expression. With 13 prototypes and seven expressions, more than one prototype can carry the same label. Prototype coordinates follow the circumplex model~\cite{russell1980circumplex}, so each basic emotion occupies a characteristic region.
Given $\mathbf{p}_t = (\tilde{z}_{t,V}, \tilde{z}_{t,A})$, we compute squared distances $d_{tj} = \|\mathbf{p}_t - \mathbf{c}_j\|^2$ and obtain fuzzy memberships by inverse-distance normalization,
$\mu_{tj} = \bigl(d_{tj}\sum_{l=1}^{P} d_{tl}^{-1}\bigr)^{-1}$ (reducing to a hard assignment when the input coincides with a prototype).
The output weight vector is the membership-weighted sum of prototype labels:
\begin{equation}
    \boldsymbol{\omega}_t = \sum_{j=1}^{P} \mu_{tj}\, \mathbf{w}_j .
    \label{eq:fkcn}
\end{equation}
Unlike nearest-neighbor hard classification, this mapping is continuous over the VA plane: a point between prototypes yields a soft blend of several expressions rather than a hard jump.

\subsubsection{Hardware Actuation and Signal Tuning}
The actuation layer fuses the seven-dimensional weight vector into servo commands.
Each basic expression has an offline-calibrated preset $\mathbf{p}^{(m)} \in [0,1]^{29}$: joints are posed for the target expression and normalized by each channel's travel limits.
At run time, the 29-dimensional target factor is a linear fusion of the presets with the FKCN weights:
\begin{equation}
    \mathbf{f}_t = \sum_{m=1}^{7} \omega_{t,m}\, \mathbf{p}^{(m)}, \qquad \mathbf{f}_t \leftarrow \mathrm{clip}(\mathbf{f}_t,\, 0,\, 1),
    \label{eq:fusion}
\end{equation}
The clip is a safeguard: with weights on the simplex and presets in $[0,1]^{29}$, the fusion already lies in $[0,1]^{29}$. The target is converted channel-wise into servo angles and sent over the serial port. We also apply exponential moving average (EMA) smoothing and frame-rate limiting to suppress per-token jitter and match servo dynamics; defaults and ablations are reported in Section~\ref{sec:exp_hardware}.

Because fusion is defined in preset-weight space, the projection and the FKCN never address the 29 servo channels. Only the seven expression presets are posed on this robot.

\section{EXPERIMENTS}
\label{sec:experiments}

We design experiments to address three core questions:
\begin{itemize}
\item \textbf{Q1. Within-Utterance Tracking:} Does the physical face reproduce the affective trajectory read token by token from the hidden states?
\item \textbf{Q2. Response Timing:} How much earlier does the face move when driven during decoding rather than after the reply?
\item \textbf{Q3. Affective Readout:} Does the coordinate read the situation, or only the words?
\end{itemize}

\subsection{Experimental Setup}

Physical evaluation uses the 29-DoF facial robot in Fig.~\ref{fig:add}, controlled by an on-board Intel N100 industrial PC (4-core CPU, no discrete GPU) over a serial port. Unless noted otherwise, all experiments use Qwen2.5-0.5B with offline affective axes and calibration parameters built from 50 English pole words per pole (Section~\ref{sec:axis_mining}).

\subsection{Main Comparison}
\label{sec:exp_hardware}

\subsubsection{Actuation Fidelity}
This experiment evaluates physical tracking of token-rate affective commands. Twenty replies are generated once by Qwen2.5-0.5B and recorded in one session under a fixed camera pose across five conditions. EmoManifold updates valence, arousal, and the facial target once per token. The four baselines share the identical reply text: Mishra et al.~\cite{mishra2023realtime} assign one of six categories to the finished reply and hold it for the turn using their original prompt; Xpress-Turn~\cite{antony2025xpress} holds one category and intensity for the turn; Xpress-Story~\cite{antony2025xpress} synthesizes speech, splits the reply at narrative or emotional turns, and plays one mapped preset per segment from its word-level timestamp; and the shuffled control replays EmoManifold's per-token commands in randomized order, preserving command amplitudes while removing temporal order. Category calls use GPT-4o-mini for the two turn-level baselines and GPT-4o for Xpress-Story.

While the robot acts out each reply, a fixed camera records the face and Py-Feat~\cite{cheong2023pyfeat} estimates frame-by-frame valence and arousal independently of our pipeline. Because Py-Feat is trained on human faces, we place its readings on the commanded scale by fitting one linear map per dimension on 40 turn-level static segments from Mishra et al.\ and Xpress-Turn, then freezing both maps across all dynamic trials. We define the commanded trace as the token-level valence and arousal trajectory sent to the actuators by EmoManifold, and score the calibrated camera readout of all five conditions against this same trace using a fixed $0.5$-s actuation lag. The primary metric is event-level directional agreement: within each 2-s sliding window, an event is flagged when commanded valence shifts by more than $0.3$ or commanded arousal by more than $0.2$, and a hit is recorded if the calibrated camera readout shifts in the same direction ($50\%$ chance level). As a secondary metric, we report the $R^2$ gain (in percentage points) over a static turn-level expression (Fig.~\ref{fig:5}(b)).

Fig.~\ref{fig:5} shows that EmoManifold tracks commanded transitions within the utterance. Directional agreement is $87.5\%$ for valence ($95\%$ CI $[76.7,96.0]$, $p<10^{-4}$) and $69.6\%$ for arousal (CI $[56.8,81.7]$, $p{=}0.001$). On valence, none of the other four conditions rises above the $50\%$ chance level ($37.2\%$--$49.4\%$). On arousal, those conditions span $34.0\%$--$58.8\%$, and at least one exceeds chance. Xpress-Story is the informative baseline: its face changes at segment boundaries within the reply, yet valence agreement stays at $49.4\%$ (CI $[31.3,66.2]$), showing that within-reply changes must land where the commanded trajectory shifts. For the $R^2$ gain, EmoManifold reaches $21.8$ percentage points in valence and $18.9$ in arousal, against $7.5$ and $5.9$ for Xpress-Story (Fig.~\ref{fig:5}(b)). A circular-shift surrogate on EmoManifold rejects chance alignment (Fisher-combined $p<10^{-23}$), and the shuffled control---which keeps command amplitudes without their order---falls back to chance. Intra-utterance tracking therefore depends on the temporal order of the commands.

\begin{table}[t]
    \caption{On-device latency (mean $\pm$ std., $n{=}20$).}
    \label{tab:latency}
    \centering
    \begin{tabular}{@{}lccc@{}}
    \toprule
    \raisebox{3pt}{System} & \shortstack{Per-token\\(ms)} & \shortstack{First update (s)\\GPT-4o-mini} & \shortstack{First update (s)\\Qwen2.5-0.5B} \\
    \midrule
    EmoManifold & $0.7{\pm}0.2$ & \multicolumn{2}{c}{$0.96{\pm}0.19$} \\
    Mishra et al.~\cite{mishra2023realtime} & --- & $10.2{\pm}4.3$ & $9.9{\pm}4.0$ \\
    Xpress-Turn~\cite{antony2025xpress} & --- & $9.9{\pm}3.8$ & $26.5{\pm}5.7$ \\
    Xpress-Story~\cite{antony2025xpress} & --- & $19.5{\pm}6.1$ & $93.5{\pm}39.6$ \\
    \bottomrule
    \end{tabular}
\end{table}

\subsubsection{On-Device End-to-End Latency}
This experiment measures end-to-end response timing and per-token control overhead on the industrial PC (Table~\ref{tab:latency}). EmoManifold delivers its first content-driven facial update in $0.96$~s, including the language-model forward pass, and the later per-token control path takes $0.7$~ms excluding that pass. Turn-level baselines finish the reply and then query a category label: with GPT-4o-mini the first update takes $10.2$~s for Mishra et al.\ and $9.9$~s for Xpress-Turn; on local Qwen2.5-0.5B, with no external call, it takes $9.9$~s and $26.5$~s. Xpress-Story's first update is that same reply generation plus its segmentation and emotion calls, $19.5$~s with GPT-4o-mini and $93.5$~s on the local model; speech synthesis is not on this clock. By reading affect directly during decoding rather than after the reply is finished, EmoManifold removes the turn-level latency bottleneck (Table~\ref{tab:design}).

\begin{table}[t]
    \caption{Comparison of facial robot interaction systems.}
    \label{tab:design}
    \centering
    \footnotesize
    \setlength{\tabcolsep}{3pt}
    \begin{tabular}{@{}lcccc@{}}
    \toprule
     & \textbf{EmoManifold} & \shortstack{Xpress-\\Story} & \shortstack{Xpress-\\Turn} & \shortstack{Mishra\\et al.} \\
    \midrule
    Source & \textbf{Hidden} & Script & Label & Label \\
    Expression & \textbf{Continuous} & Discrete & Discrete & Discrete \\
    Update & \textbf{Token} & Keyframe & Turn & Turn \\
    Latency & \textbf{0.96\,s} & $93.5$\,s & $26.5$\,s & $9.9$\,s \\
    \bottomrule
    \end{tabular}
\end{table}

\subsection{Reliability of the Affective Subspace}
\label{sec:exp_subspace}

We next examine whether the mined VAD axes recover human lexicon ratings and maintain a consistent orthogonal geometry across models and languages. We evaluate six language models (Table~\ref{tab:e2_cross_model_spearman}) on 300 held-out single-token words from the NRC-VAD lexicon (100 per dimension, grouped by their most salient dimension), computing the Spearman correlation $\rho$ between projected unembedding scores and ground-truth lexicon ratings.

\begin{figure}[t]
    \centering
    \includegraphics[width=\columnwidth]{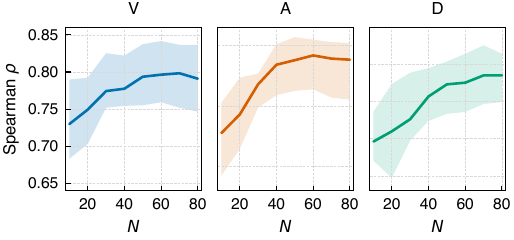}
    \caption{Word-level Spearman correlation as a function of the number of pole words $N$. Solid lines show the mean across six models; shaded bands show the min--max range. All three dimensions saturate beyond $N \approx 50$.}
    \label{fig:2}
\end{figure}

Axis quality depends on the pole-word count $N$. Fig.~\ref{fig:2} shows $\rho$ across the six models as $N$ varies from 10 to 80: all three dimensions are unstable at small $N$, improve as $N$ grows, and saturate beyond $N \approx 50$, which we adopt by default.

\begin{table}[t]
    \caption{Cross-model word-level Spearman correlation ($N{=}50$).}
    \label{tab:e2_cross_model_spearman}
    \centering
    \begin{tabular}{lccc}
    \toprule
    Model & $\rho_V$ & $\rho_A$ & $\rho_D$ \\
    \midrule
    Gemma-2-2B & 0.802 & 0.650 & 0.598 \\
    Llama-3.2-3B & 0.800 & 0.828 & 0.708 \\
    Mistral-7B-v0.3 & 0.794 & 0.788 & 0.567 \\
    Qwen2.5-7B & 0.755 & 0.792 & 0.648 \\
    Llama-3.1-8B & 0.773 & 0.788 & 0.697 \\
    Qwen2.5-0.5B & 0.837 & 0.665 & 0.663 \\
    \bottomrule
    \end{tabular}
    \end{table}

Table~\ref{tab:e2_cross_model_spearman} reports per-dimension correlations at $N{=}50$, with cross-dimension means falling between $0.68$ and $0.78$. The deployed Qwen2.5-0.5B model reaches a valence correlation of $0.837$, matching models up to 8B. Rebuilding the axes from Chinese pole words on Qwen2.5-0.5B yields $\rho_V = 0.823$, $\rho_A = 0.803$, and $\rho_D = 0.702$, confirming a linearly readable affective structure in both languages.

\begin{figure}[t]
    \centering
    \includegraphics[width=\columnwidth]{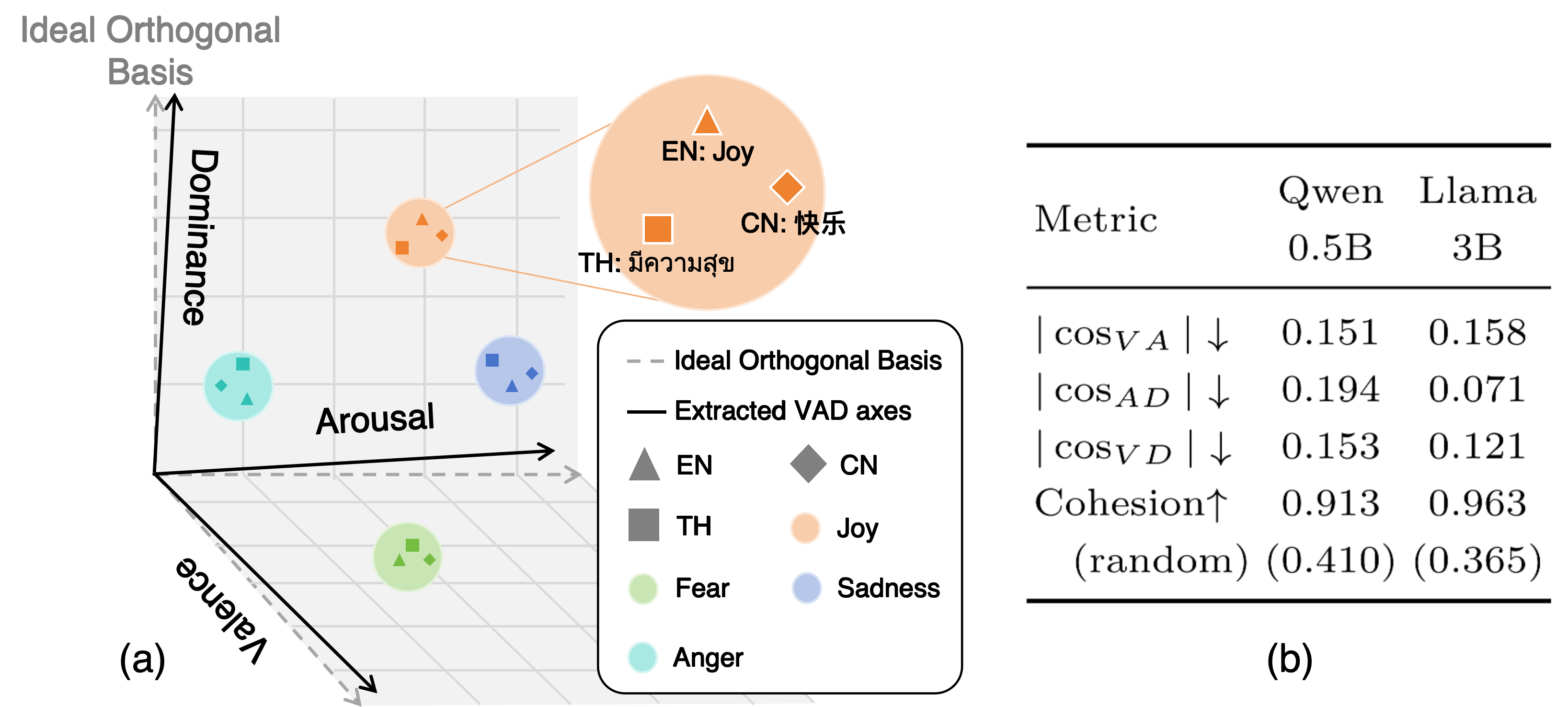}
    \caption{Cross-lingual affective geometry ($|\cos|$, cohesion, and the random baseline are defined in the text). \textbf{(a)}~Schematic of cross-lingual emotion word clusters (illustrative, not measured): solid lines are the three mined affective axes; dashed lines are an idealized orthogonal basis. \textbf{(b)}~Measured cohesion after each backbone builds its axes from English pole words ($N{=}50$). Parenthesized numbers are that model's random baseline.}
    \label{fig:3}
\end{figure}

To check axis orthogonality and cross-lingual consistency, we build axes from English pole words ($N{=}50$) on Qwen2.5-0.5B and Llama-3.2-3B, then project five emotion clusters (fear, comfort, dominance, strength, calm) containing single-token synonyms in five languages per cluster (English, Chinese, and three additional languages). Cluster cohesion is measured as the mean pairwise cosine similarity within each cluster and compared against a random three-dimensional orthogonal basis (100 trials per cluster).

Fig.~\ref{fig:3}(a) illustrates this geometry, and Fig.~\ref{fig:3}(b) reports the metrics for English-built axes: on Qwen2.5-0.5B, $|\cos|\le0.194$ with a cohesion of $0.913$ (random: $0.410$); on Llama-3.2-3B, $|\cos|\le0.158$ with a cohesion of $0.963$ (random: $0.365$). Axes rebuilt from Chinese pole words remain in the same range ($|\cos|\le0.220$ and cohesion $0.898$ on Qwen; $|\cos|\le0.282$ and cohesion $0.965$ on Llama), showing that the subspace geometry holds across both backbones and source languages.

\subsection{Context Sensitivity}
\label{sec:exp_sarcasm}

\begin{figure}[t]
    \centering
    \includegraphics[width=\columnwidth]{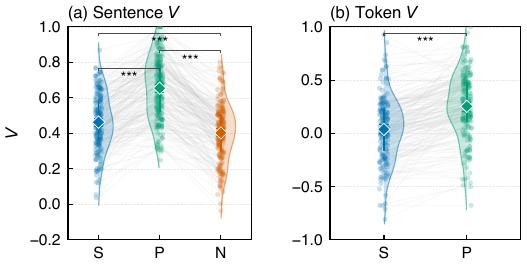}
    \caption{Contextual-triplet sarcasm stress test ($n{=}200$, Qwen2.5-0.5B; triplet construction in the text).
    \textbf{(a)}~Sentence-level projected valence $V$ for S (sarcastic), P (sincere-positive), and N (literal-negative). Raincloud plots combine a half-violin (kernel density), a box plot (interquartile range, IQR), and per-triplet scatter points; diamonds are bootstrap means $\pm$95\% CI, gray lines connect matched triplets.
    \textbf{(b)}~Token-level valence at the primary anchored positive keyword, S versus P ($n{=}194$).}
    \label{fig:4}
  \end{figure}

This experiment tests whether the readout reflects situational context rather than static word priors, using sarcasm where positive wording carries negative intent~\cite{abufarha2022isarcasm}. We construct 200 contextual triplets from iSarcasmEval SubTask~C~\cite{abufarha2022isarcasm}, formatting each item as one sentence of observable context followed by a quoted utterance: S (sarcastic) pairs an adverse context with a positive quotation; P (sincere-positive) pairs a favorable context with the word-for-word identical positive quotation; and N (literal-negative) pairs the adverse context with a literally negative quotation. Each triplet is annotated with 1--3 anchored positive keywords (e.g., ``well'') for token-level evaluation.

We read out sentence-level valence at the final token and token-level valence at each anchored keyword. Across the 200 triplets (Fig.~\ref{fig:4}(a)), sentence-level valence for S and N closely overlaps ($81.5\%$ of pairs differ by less than $0.15$), whereas S drops significantly below P despite their identical quotations ($\Delta V=-0.19$; Wilcoxon, $n{=}200$, $p<10^{-11}$).

To isolate single-token contextual shifts, Fig.~\ref{fig:4}(b) compares the primary anchored keyword appearing in the exact same position and surface form in S and P. Under sarcasm, keyword valence drops by $0.22$ on average ($-0.20$ across all co-occurring keywords), while arousal and dominance shift by $+0.08$ and $-0.07$, matching the valence-driven evaluation shift expected in verbal irony~\cite{leggitt2000emotional}. Repeating the test on Llama-3.2-3B yields the same directional shifts ($\Delta V(S{-}P)=-0.20$ at the sentence level and $-0.16$ at the primary keyword), confirming that the readout reflects situational context rather than static word priors.


\subsection{Robustness}
This experiment tests whether the display pipeline preserves within-utterance command order across smoothing rates and prototype counts. We replay the pipeline on the 20 recorded command trajectories, read each frame-rate output back onto the VA plane, and score it against the token-level command using the same event thresholds and windows. At the default setting ($\alpha{=}0.6$, $P{=}13$), replay gives $84.4\%$ valence and $67.7\%$ arousal agreement, closely matching the camera measurements ($87.5\%$ and $69.6\%$) and validating replay as a proxy for physical execution. Time-shuffled commands passed through the same pipeline fall to $44\%$--$48\%$ valence agreement ($40\%$--$45\%$ arousal), confirming that the metric remains sensitive to command order.

We then sweep nine settings formed by crossing three smoothing rates ($\alpha\in\{0.3,0.6,0.9\}$) with three prototype sets---the default $P{=}13$ and two $P{=}7$ constructions that each retain a different prototype per emotion class---under the same per-frame rate limit. Across all nine settings, valence agreement stays within $75\%$--$87\%$ and arousal agreement within $68\%$--$74\%$, well above the shuffled range, showing that temporal command order survives the display pipeline across parameter choices.

\subsection{Human Study}
\label{sec:exp_human}
To evaluate human perception, we conduct two silent-video questionnaires with 20 participants (10 items each) assessing expression recognition and dynamic preference. Each item displays the full sentence text before playing the unsubtitled video; participants gave informed consent, and all responses were anonymous.

In the recognition task, participants classify each video into one of the seven basic expressions, compared against hand-tuned static references recorded at the same camera pose. EmoManifold achieves $85.5\%$ accuracy versus $90.5\%$ for the hand-tuned references (Wilcoxon $p{=}0.013$), with both showing identical confusion patterns between neighboring pairs (fear--surprise and sadness--disgust) inherent to the robot's facial mechanics.

In the preference task, participants make a three-way choice on sentences containing an intra-utterance emotional shift, selecting which video best matches the emotional progression: EmoManifold's dynamic expression, a static expression held for the whole sentence, or the shuffled-order control. EmoManifold is chosen on $83.5\%$ of trials, compared with $11.5\%$ for the shuffled control and $5.0\%$ for the static expression, with every participant preferring EmoManifold over both alternatives (one-sided Wilcoxon over participants, $p<10^{-6}$). Expression recognition stays within five percentage points of the hand-tuned references, and observers match the token-level video to emotional shifts inside the sentence.

\section{DISCUSSION}
\label{sec:discussion}

Collectively, our offline readout and physical hardware evaluations demonstrate that continuous facial expressions can be driven directly from LLM hidden states. The projected coordinates correlate consistently with lexicon ratings across six language models in both English and Chinese, and respond to shifts in situational context even when surface wording is held constant. When mapped through the FKCN and expression presets onto the physical robot, these trajectories are reliably recovered by external camera tracking and preferred by human observers over both static turn-level expressions and time-shuffled controls. Compared with discrete emotion labels inferred only after a reply completes, the principal advantage is temporal: affective dynamics are rendered concurrently with language generation rather than deferred until the end of the utterance.

First, coupling this affective channel to audible speech playback remains open. The evaluations establish expression--content alignment at token rate. The human studies use silent video, and the current system does not drive audio-synchronized lip motion. Because mouth degrees of freedom must serve both articulation and expression, a naive superposition of the two can cause mechanical interference. A direct next step is to apply the same hidden-state readout to open-source speech-native models and project affect along the audio-token stream, so facial updates share the playback clock while lip articulation runs as a complementary controller on that timeline.

\addtolength{\textheight}{-4.5cm}

Second, the face hardware and the interaction policy both constrain what can be shown. Discrimination among neighboring emotions is bounded by the robot's physical DoFs and preset expression basis. More critically, directly externalizing an internal affective state can be socially inappropriate: natural human interaction often requires suppressing, softening, or delaying expression, whereas the current pipeline renders the signal immediately upon readout. We therefore plan a lightweight, optional regulation stage between projection and actuation---gain limiting, temporal hold, and an explicit off switch---so deployment can choose when the internal signal is shown, softened, or suppressed.

Third, the physical and human studies remain modest in scale across corpus size, participant count, and tested backbones. The closed-loop physical actuation (Section~\ref{sec:exp_hardware}) and the human study (Section~\ref{sec:exp_human}) use only Qwen2.5-0.5B. Because the subspace and context-sensitivity experiments already show comparable geometry and sensitivity on Llama-3.2-3B, we concentrated hardware and human-evaluation resources on the deployment model. Repeating the full closed-loop protocol on a second backbone, with a larger participant pool, is the direct next measurement step along the same evaluation pipeline.

\section{CONCLUSION}
\label{sec:conclusion}

This paper shows that a facial robot need not rely on a discrete emotion label: what a model will express in words or audio is already computed in its hidden states before the output is rendered. Emo-H2A takes this as its premise, locating the relevant hidden states, extracting the affective signal---whether by subspace projection, as in EmoManifold, or by a learned network---and running actuation in parallel with generation rather than after it. Continuous VAD coordinates are read from an LLM's hidden states during streaming decoding, with the valence--arousal plane driving physical expression at token granularity. Across experiments, the training-free pipeline yields a context-sensitive readout ($\Delta V{=}-0.19$ under sarcasm), tracks commanded intra-utterance transitions on a facial robot ($87.5\%$ valence and $69.6\%$ arousal directional agreement, against near-chance baselines and with $83.5\%$ human preference), and reduces on-board first-update latency to $0.96$~s, compared with $9.9$--$93.5$~s for local baselines. These results offer a practical route beyond turn-level control toward fast, fine-grained, and verifiable embodied affective interaction.

\bibliographystyle{IEEEtran}
\bibliography{references_en}

\end{document}